%% file: acl_latex.tex
\pdfoutput=1
\documentclass[11pt]{article}
\usepackage{acl}
\usepackage{times}
\usepackage{latexsym}
\usepackage[T1]{fontenc}
\usepackage[utf8]{inputenc}
\usepackage{microtype}
\usepackage{inconsolata}
\usepackage{graphicx}
\usepackage{xspace}
\usepackage{graphicx}
\usepackage{amsmath}
\usepackage{amsfonts}
\usepackage{float}
\usepackage{booktabs}
\usepackage{wrapfig}
\usepackage{subcaption}
\usepackage{hyperref}
\usepackage{xurl}
\newcommand{\ourmethod}{\textsc{SAFR}\xspace}
\newcommand{\ourfullmethod}{\textbf{S}uperposition-\textbf{A}ware \textbf{F}eature \textbf{R}egularization\xspace}
\newcommand{\eval}{\textsc{SRS}\xspace}

\title{SAFR: Neuron Redistribution for Interpretability}

\author{Ruidi Chang \quad Chunyuan Deng \quad Hanjie Chen \\
Department of Computer Science \\
Rice University\\
\texttt{\{ruidi.chang, hanjie\}@rice.edu} \\
}

\begin{document}
\maketitle
\input{sections/Abstract}
\input{figures/Workflow}
\input{sections/Introduction}
\input{sections/Preliminaries}
\input{tables/Main_results}
\input{sections/Methodology}
\input{sections/Experiment}
\input{figures/Results}
\input{sections/Results}
\input{sections/RelatedWork}
\input{sections/Conclusion}
\input{sections/Limitation}
\input{sections/Ethics}
\input{sections/Acknowledgment}

\bibliography{custom}

\appendix
\input{sections/Vmask}
\input{sections/Statistics}
\input{sections/Visualization}
\input{sections/Ablation}

\end{document}

%% file: sections/Abstract.tex
\begin{abstract}
Superposition refers to encoding representations of multiple features within a single neuron, which is common in deep neural networks. This property allows neurons to combine and represent multiple features, enabling the model to capture intricate information and handle complex tasks. Despite promising performance, the model's interpretability has been diminished. This paper presents a novel approach to enhance model interpretability by regularizing feature superposition. We introduce \ourmethod,\footnote{\ourmethod: \ourfullmethod} which simply applies regularizations to the loss function to promote monosemantic representations for important tokens while encouraging polysemanticity for correlated token pairs, where important tokens and correlated token pairs are identified via VMASK \cite{chen-ji-2020-learning} and attention weights respectively. We evaluate \ourmethod with a transformer model on two classification tasks. Experiments demonstrate the effectiveness of \ourmethod in improving model interpretability without compromising prediction performance. Besides, \ourmethod provides explanations by visualizing the neuron allocation within the intermediate layers.\footnote{The code can be found in \href{https://github.com/chili-lab/SAFR}{\url{https://github.com/chili-lab/SAFR}}.}

\end{abstract}

%% file: figures/Workflow.tex
\begin{figure*}[t]
\centering
  \includegraphics[width=0.9\linewidth]{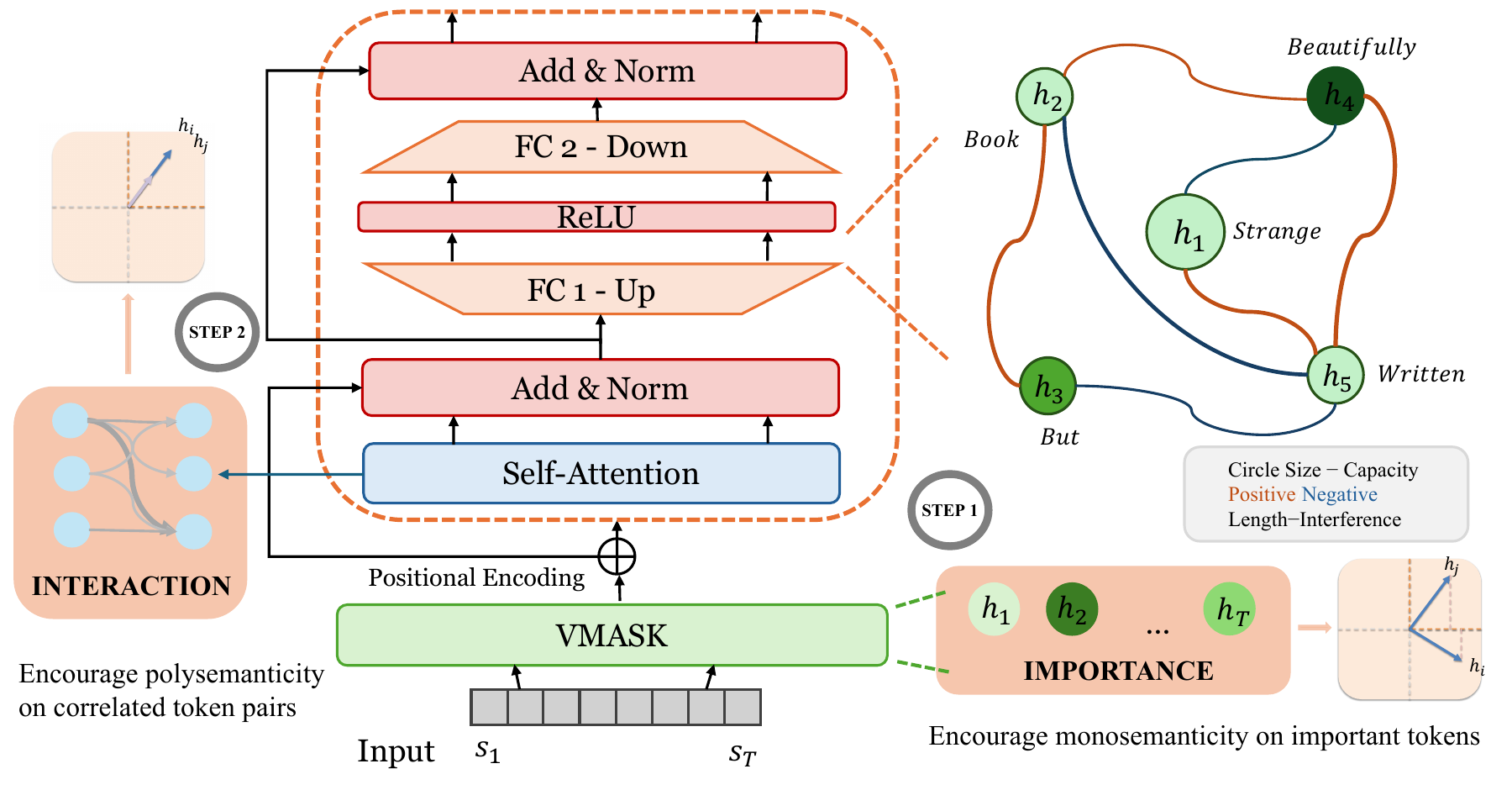}
  \caption{\textbf{Basic structure of \ourmethod}.
i) Promote monosemanticity for important tokens after the embedding layers.  
ii) Leverage the attention mechanism to enhance polysemanticity among correlated token pairs.  
   }
  \label{fig:workflow}
\end{figure*}

%% file: sections/Introduction.tex
\section{Introduction}

Individual neurons in neural networks can represent multiple features from the input. This phenomenon, known as superposition, improves the model's ability to capture intricate relationships between features~\cite{olah2020zoom}, while also complicating the understanding of the underlying processes behind the model's decision-making~\cite{elhage2022toy}. Facing these challenges, recent research like sparse autoencoders (SAEs)~\cite{huben2024sparse}  artificially decomposes the activation space into a sparse vector space through auxiliary networks. While SAEs provide a method to \textit{interpret} features through combinations of sparse activations, there is still a lack of sufficient research on \textit{controlling} neuron distribution for interpretability.

In this paper, we ask the question: \textit{Can we enhance model interpretability by explicitly controlling the distribution of features across neurons?} An intuitive approach is to encourage monosemantic neurons by regulating activations \cite{elhage2022toy, bricken2023monosemanticity, wang2024learningemergencestudyproactively}. However, focusing solely on monosemanticity may limit the model's ability to capture feature interactions, potentially hindering overall performance.

To address this challenge, we propose a novel method called \ourfullmethod (SAFR) to enhance model interpretability by strategically redistributing neurons through a modified loss function, approached from two perspectives. As illustrated in Figure \ref{fig:workflow}, our framework incorporates regularization techniques aimed at promoting monosemantic representations for important tokens, while simultaneously fostering polysemanticity among correlated token pairs. The identification of important tokens is achieved via a variational inference network, adapted from VMASK~\cite{chen-ji-2020-learning}. Additionally, correlated token pairs are identified based on attention weights to foster polysemantic representation.

We evaluate the effectiveness of \ourmethod with a transformer model on the SST-2 \cite{socher-etal-2013-recursive} and IMDB \cite{maas-etal-2011-learning} datasets. When the top 30\% of words identified by \ourmethod are removed, we observe a significant drop in test accuracy—17.21\% on SST-2 and 28.48\% on IMDB. As we gradually remove additional words, accuracy continues to decline in a consistent manner. To further substantiate these findings, we visualize the neuron allocation within the FFN layers of a transformer block. Our experimental results demonstrate that \ourmethod can effectively redistribute features across neurons while preserving a reasonable degree of polysemanticity, significantly improving interpretability while maintaining prediction performance at comparable levels.

%% file: sections/Preliminaries.tex
\section{Preliminaries}

Given an input sequence of \( T \) tokens \( S = \langle s_1, \ldots, s_T \rangle \) and a neural network model $f(\cdot)$ with \( L \) layers, let \( h^\ell_i \) denote the hidden representation obtained at layer \( \ell \in [1, \ldots, L] \) and token position \( i \in [1, \ldots, T] \). Following previous work ~\cite{scherlis2022polysemanticity, elhage2022toy}, which defines \textit{interference}, \textit{polysemanticity}, and \textit{capacity} based on features, we extend these definitions to hidden representations in our analysis.

\paragraph{Interference}
The \textit{interference} (\( I \)) measures the overlap or similarity between two representations at the same layer:
\begin{equation}
  \textstyle I_{i, j}^\ell = h^\ell_i  \cdot h^\ell_j
\end{equation}
Interference quantifies how much the hidden representations interfere with each other. Higher interference indicates greater overlap, suggesting the tokens share representational dimensions.
\paragraph{Polysemanticity}
The \textit{polysemanticity} (\( P \)) describes the extent to which a single hidden representation captures information from multiple tokens:
\begin{equation}
\textstyle P_i^\ell = \sum_{j \neq i} \left( \hat{h}^\ell_i \cdot h^\ell_j \right)^2   
\end{equation}
Here, \( \hat{h}^\ell_i \) denotes the normalized representation of \( h^\ell_i \) (divided by its own magnitude). A high polysemanticity value indicates that a single direction in the representation space is ``polysemantic'', meaning it simultaneously represents information from multiple tokens.
\paragraph{Capacity}
The \textit{capacity} (\( C \)) quantifies how much of a hidden representation direction is dedicated to representing token \( i \):
\begin{equation}
\textstyle C_i^\ell = \frac{(h^\ell_i  \cdot h^\ell_i )^2}{\sum_{j} (h^\ell_i \cdot h^\ell_j )^2}  
\end{equation}
where \( 0 \leq C_i^\ell \leq 1 \) and \( 1 \leq C^\ell \leq T \) with \(C^\ell = \sum_{i=1}^{T} C_i^\ell\) and \(T\) denotes sequence length. A higher capacity value indicates that the representation at position \( i \) is more focused on representing the $i$-th token in the input.

%% file: tables/Main_results.tex
\begin{table*}[ht]
    \centering
    \small
    \renewcommand{\arraystretch}{0.8}
    \resizebox{\textwidth}{!}{%
    \begin{tabular}{lcccccccc}
        \toprule
        & \multicolumn{4}{c}{SST-2} & \multicolumn{4}{c}{IMDB} \\
        \cmidrule(lr){2-5} \cmidrule(lr){6-9}
        Model & $\text{Acc}_{S} (\%)$ & $\text{Acc}_{\tilde{S}^{(r)}} (\%)$ & $\text{Acc}_{\tilde{S}^{(k)}} (\%)$ & \eval & $\text{Acc}_{S} (\%)$ & $\text{Acc}_{\tilde{S}^{(r)}} (\%)$ & $\text{Acc}_{\tilde{S}^{(k)}} (\%)$ & \eval \\
        \midrule
        Baseline & 70.21 & 67.12 & 66.21 & 4.00 & 80.14 & 77.11 & 76.60 & 3.54 \\
        $\lambda_{Imp} = 0, \lambda_{Inter} = 0$ & 72.56 & 69.47 & 64.44 & 8.12 & 78.43 & 76.10 & 73.56 & 4.87\\
        \ourmethod & 72.96 & 70.61 & 55.75 & \textbf{17.21} & 78.45 & 75.05 & 49.97 & \textbf{28.48}\\
        \bottomrule
    \end{tabular}}
    \caption{Evaluation on SST-2 and IMDB datasets, k = 30. 
    $\text{Acc}_{S}$ denotes prediction accuracy for original test dataset. \(\text{Acc}_{\tilde{S}^{(r)}}\) and \(\text{Acc}_{\tilde{S}^{(k)}}\) denote prediction accuracy after randomly deleting \( k\% \) of tokens and deleting \( k\% \) based on capacity, respectively. \ourmethod achieves a \eval score of 17.21 for SST-2 and 28.48 for IMDB, outperforming the baseline model, indicating an improvement on model interpretability. The optimal parameter settings for SAFR are $\lambda_{Imp} = 0.1, \lambda_{Inter} = 0.1$ for SST-2 and $\lambda_{Imp} = 0.1, \lambda_{Inter} = 1$ for IMDB.
}
    \label{tab:main_results}
\end{table*}

%% file: sections/Methodology.tex
\section{Methodology}

\ourmethod enhances model interpretability by strategically redistributing neurons through a superposition regularization strategy. By promoting a well-structured neuron distribution that balances importance and interaction, it makes token representations more meaningful.

The baseline is defined using the original model with cross-entropy loss \(
\mathcal{L}_{\mathrm{CE}} = -\frac{1}{N}\sum_{n=1}^{N} \sum_{g=1}^{G} y_n^{g} \log p_n^{g}
\)
, where $N$ is the number of samples in the dataset, and $G$ is the number of classes in text classification, 
$y_n^{g}$ is an indicator that equals 1 if sample $n$ belongs to class $g$ (and 0 otherwise), $p_n^{g}$ is the predicted probability for sample $n$ being of class $g$. To improve model interpretability in a constrained representational space, we propose a two-part regularization strategy: one that promotes monosemantic representations for important tokens, and the other encourages polysemanticity for correlated token pairs. This approach enables the model to effectively allocate its representational resources, as shown in Figure \ref{fig:workflow}.

The proposed loss function integrates these regularization terms accordingly:
\begin{align*}
\mathcal{L} &= \mathcal{L}_{\text{CE}} + \lambda_{Imp}\!\cdot\!\mathcal{L}_{\text{Importance}} + \lambda_{Inter}\!\cdot\!\mathcal{L}_{\text{Interaction}}
\end{align*}
where \(\lambda_{Imp}\) controls the importance loss term and \(\lambda_{Inter}\) controls the interaction loss term.

\paragraph{Importance-Based Regularization}
We apply the VMASK~\cite{chen-ji-2020-learning} between the embedding layer and the positional encoder to select important tokens. A detailed introduction to VMASK is provided in Appendix~\ref{vmask}. To encourage monosemanticity for important tokens, we introduce a regularization term \(\mathcal{L}_{\text{Importance}}=\frac{1}{T}\sum_{i=1}^{T} \sqrt{P_{i}^{V}/{E}}\), where \(E\) represents the embedding dimension and \(P_{i}^{V}\) denotes the polysemanticity for the hidden representation of the $i$-th token after the VMASK layer. This regularization penalizes important tokens with high polysemanticity.

\paragraph{Interaction-Based Regularization}
We leverage the attention mechanism and employ the attention weights to identify correlated tokens. The $\alpha$-th self-attention head is described as follows:
\begin{center}
\(
A_{\alpha} = \text{softmax}\left(\frac{Q_{\alpha} K_{\alpha}^T}{\sqrt{d_k}}\right)
\)
\end{center}
where \( \alpha \in [1, \ldots, M]\), \(Q_{\alpha} = \mathbf{X}W^Q_{\alpha}\) and \(\quad K_{\alpha} = \mathbf{X}W^K_{\alpha}\), with \(Q_{\alpha}, K_{\alpha} \in \mathbb{R}^{T \times d_k}\), \(d_k\) denotes the dimension of the key and query vectors. \(\mathbf{X}= E_{\text{pos}}(S') \in \mathbb{R}^{T \times E}\)
denotes the input matrix to the attention layer, where \(E_{\text{pos}}(S')\) is the positional encoding applied to the output \(S'\) from the VMASK layer. The score \(A_{\alpha(i,j)}\) indicates how much attention token \(i\) places on token \(j\).

To encourage highly correlated token pairs to exhibit high polysemanticity, we introduce a loss term \(\mathcal{L}_{\text{Interaction}}=\sum_{\alpha} \sum_{i, j} \frac{1}{T^2} A_{\alpha(i,j)} \cdot (1 - I_{i,j}^{A_{\alpha}})\), where \(I_{i,j}^{A_{\alpha}}\) is the \(Interference\) of the attention weights matrix for the $\alpha$-th attention head. This loss term penalizes highly correlated tokens that exhibit low interference values.

\paragraph{Proposed Loss Function}
The loss term is now defined as:
\begin{align*}
\mathcal{L} &= \mathcal{L}_{\text{CE}} + \lambda_{Imp} \cdot \mathcal{L}_{\text{Imp}} + \lambda_{Inter} \cdot \mathcal{L}_{\text{Inter}} \\
&= \mathcal{L}_{\text{CE}} 
+ \lambda_{Imp}\!\cdot\!\frac{1}{NT} \sum_{n=1}^{N} \sum_{i=1}^{T} \sqrt{\frac{P_{i}^{V}}{E}} \\
&\quad + \lambda_{Inter}\!\cdot\!\frac{1}{NMT^2}\!\sum_{n=1}^{N}\!\sum_{\alpha=1}^{M}\!\sum_{i, j}\!
A_{\alpha(i,j)}\!(1\!-\!I_{i,j}^{A_{\alpha}})
\end{align*}

%% file: sections/Experiment.tex
\section{Experimental Setup}

The proposed method is evaluated on two classification tasks using a standard transformer model.

\paragraph{Datasets} We adopt two benchmark datasets: Stanford Sentiment Treebank binary version SST-2 \cite{socher-etal-2013-recursive} and movie reviews IMDB \cite{maas-etal-2011-learning}. Table~\ref{tab:data_statistics} in Appendix \ref{statistics} presents the dataset statistics.

\paragraph{Model} We use a typical transformer architecture with a single layer, following the standard setup \cite{vaswani2017attention}. This includes the complete transformer framework with its attention mechanism and positional encoding. In the multi-layer perceptron (MLP) section, we use two fully connected layers: we first expand the dimensionality by a factor of four, apply a ReLU activation, then reduce it back by the same factor, aligning with the commonly used configuration in transformer models. The model uses random embeddings to avoid the influence of pre-trained embedding information. Table~\ref{tab:model_statistics} in Appendix \ref{statistics} presents the model statistics.

\paragraph{Baseline} Since our objective is to investigate how regularization can modify neuron resource allocation to enhance interpretability, we employ a standard Transformer model, without any modifications, as the baseline for comparison.

\paragraph{Evaluation}
To evaluate our method, we define an evaluation metric called \textbf{S}uperposition \textbf{R}egularization \textbf{S}core (\textbf{SRS}). By deleting the top \( k\% \) of tokens based on capacity, \textbf{SRS} calculates the average change in the prediction accuracy over all test data as follows:
\[
SRS(k)\!=\!\frac{1}{N}\!
(\sum_{S=1}^{N}\!1\!\cdot\!(\hat{y}_S = y)\!-\!\sum_{S=1}^{N}\!1\!\cdot\!(\hat{y}_{\tilde{S}^{(k)}}\!=\!y))
\label{equation:E}
\]
where \( \tilde{S}^{(k)} \) is constructed by dropping the \( k\% \) top-scored tokens identified by \ourmethod from \( S \). The SRS measures how effectively the model arranges neurons. The SRS metric quantifies the alignment between neuron allocations and the semantic significance of tokens. By assessing the structure of neuron allocation, SRS provides insights into the interpretability of the model's internal encoding of information. Higher SRS values indicate that the removed words were highly important, signifying stronger superposition regularization.

%% file: figures/Results.tex
\begin{figure*}[t]
    \centering
     \vspace{-0.5cm}
    \resizebox{\textwidth}{!}{%
        \begin{subfigure}[b]{0.3\textwidth}
            \includegraphics[width=\linewidth]{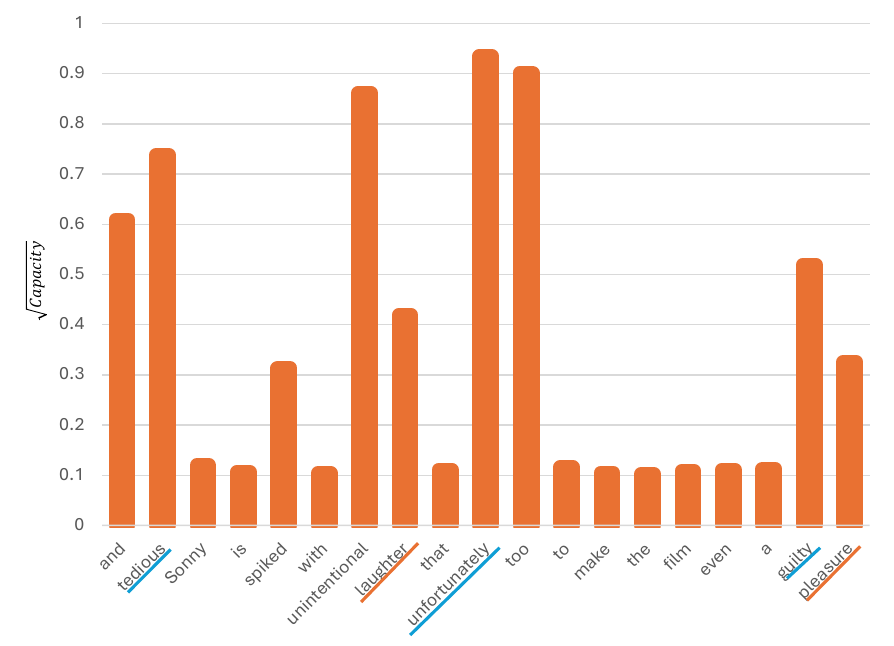}
            \caption{Capacity Barchart of FC1.}
            \label{fig:capacity}
        \end{subfigure}
        \hfill
        \begin{subfigure}[b]{0.32\textwidth}
            \includegraphics[width=\linewidth]{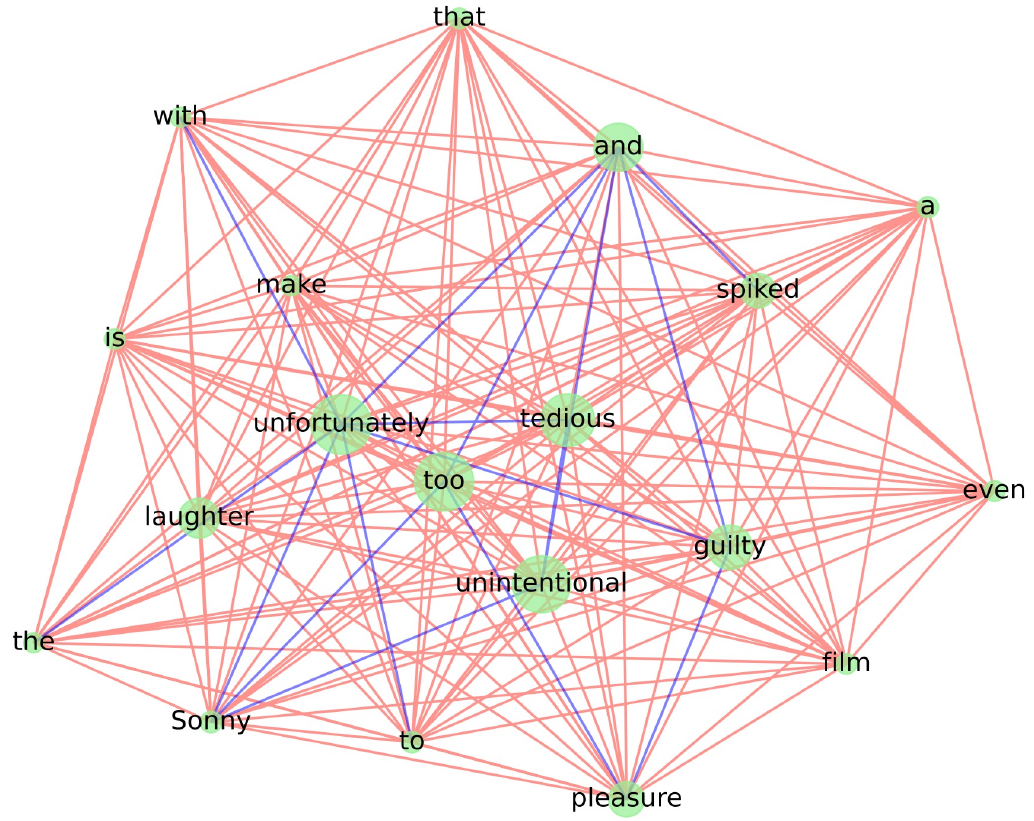}
            \caption{Visualization of FC1.}
            \label{fig:visualization}
        \end{subfigure}
        \hfill
        \begin{subfigure}[b]{0.28\textwidth}
            \includegraphics[width=\linewidth]{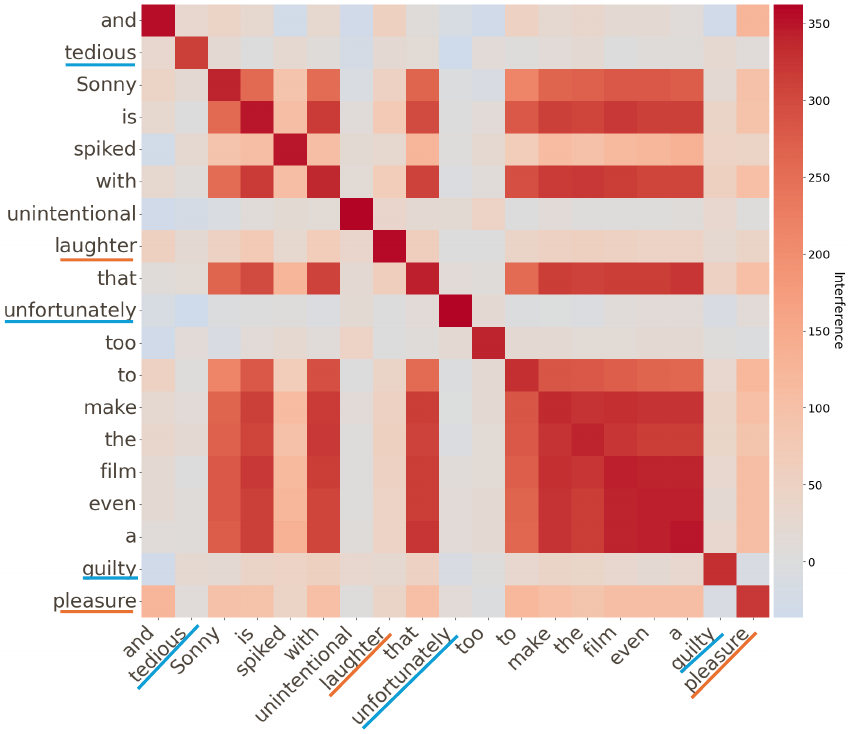}
            \caption{Interference Heatmap of FC1.}
            \label{fig:interference}
        \end{subfigure}
    }
    \caption{(a) Important tokens exhibit higher capacity. (b) Circle size represents capacity, with larger circles indicating greater capacity. Red lines denote positive correlations, blue lines indicate negative correlations, and shorter lines indicate stronger correlations. (c) Important tokens demonstrate lower polysemanticity, while correlated token pair exhibit relatively higher interference.}
    \label{fig:results}
\end{figure*}

%% file: sections/Results.tex
\section{Results and Analysis}

\paragraph{\ourmethod Improves Interpretability.}
After applying \ourmethod, the first fully connected layer redistributes neurons, making \textit{Capacity} interpretable by jointly capturing interference and polysemanticity, thereby reflecting both interaction and importance, as validated by SRS evaluation. 
Table~\ref{tab:main_results} shows that \ourmethod achieves \eval scores of 17.21 and 28.48 on the SST-2 and IMDB datasets, outperforming the baseline and demonstrating improved interpretability. 
Notably, even without regularization ($\lambda_{Imp}=0$, $\lambda_{Inter}=0$), 
the VMASK layer before the Transformer block \input{tables/Ablation}enhances \eval scores, indicating its positive effect on neuronal allocation. 
Table~\ref{tab:ablation} presents an ablation study of $\lambda_{Imp}$ and $\lambda_{Inter}$, highlighting two trade-offs: excessive $\lambda_{Imp}$ may compromise the model's ability to capture feature interactions, while high $\lambda_{Inter}$ without balancing feature importance can reduce attention to critical tokens. Additional ablation studies are in Appendix~\ref{Regularization Hyperparameter Tuning}.

\paragraph{Sensitivity to $k$ Selection.}
Figure~\ref{fig:sensitivity} illustrates the effect of token removal based on capacity. The k\% top-scored tokens are directly removed from the original text to ensure that the evaluation reflects the model’s ability to perform with reduced input information. As tokens are gradually removed, accuracy declines consistently. The greater decline in our model compared to the baselines suggests better interpretability.

\input{figures/Sensitivity}

\paragraph{Average Capacity Across the Vocabulary.}
Table~\ref{tab:capacity} presents an analysis of the average capacity per token status (important or not w.r.t VMASK) on the entire test set. Tokens are categorized based on VMASK scores, with the top 30\% identified as important and the rest 70\% as less important. These results demonstrate SAFR’s effectiveness in prioritizing and allocating greater representational capacity to task-relevant tokens. This strategic allocation improves the model's clarity in decision-making.

\input{tables/Capacity}

\paragraph{Neuron Allocation Across Layers.}
We analyzed neuron allocation across layers, with visualizations in Appendix~\ref{Neuron Allocation Accross Layers}. Neurons in the embedding layer are randomly distributed, failing to effectively capture their relative importance or inter-neuronal interactions. The VMASK layer begins to identify important tokens at a global level, yet it lacks the capability to analyze token-to-token interactions. While the attention layer demonstrates proficiency in capturing inter-token relationships, the interpretation of token importance remains debated~\cite{jain2019attention}. The second fully connected layer compresses information into a lower-dimensional space and tend to allocate neuron capacity in a uniform manner.
The expansion step in the feedforward network (FC1 layer in our model) allows the network to capture high-dimensional representations of the data. Our observation of this layer reveals that important tokens exhibit greater monosemanticity, and correlated token pairs demonstrate higher interference, illustrated in Figure \ref{fig:capacity} and \ref {fig:interference}.

\paragraph{Visualization of FC1}
Figure~\ref{fig:visualization} illustrates the enhanced interpretability achieved after applying \ourmethod. In the visualization, the size of the circles represents the capacity of individual tokens, where larger circles indicate greater capacity. The red and blue lines depict interference between tokens, with red lines corresponding to positive correlations and blue lines indicating negative correlations. The length of these lines reflects the strength of the correlation. 
This visualization illustrates neuron distribution and token relationships, demonstrating \ourmethod's effectiveness in enhancing feedforward layer interpretability. By visualizing these dynamics, the figure highlights key insights into token importance and their interactions within the model.

%% file: tables/Ablation.tex
\begin{table}[h]
    \centering
    \resizebox{\columnwidth}{!}{%
    \begin{tabular}{lcccccccc}
        \toprule
        & \multicolumn{4}{c}{SST-2} \\
        \cmidrule(lr){2-5} \cmidrule(lr){6-9}
        Model & $\text{Acc}_{S} (\%)$ & $\text{Acc}_{\tilde{S}^{(r)}} (\%)$ & $\text{Acc}_{\tilde{S}^{(k)}} (\%)$ & \eval \\
        \midrule
        Baseline & 70.21 & 67.12 & 66.21 & 4.00 \\
        $\lambda_{Imp}=0$, $\lambda_{Inter}=0$    & 72.56 & 69.47 & 64.44 & 8.12 \\
        $\lambda_{Imp}=0.1$, $\lambda_{Inter}=0.1$     & 72.96 & 70.61 & 55.75 & \textbf{17.21} \\
        $\lambda_{Imp}=0.1$, $\lambda_{Inter}=1$     & 72.67 & 69.18 & 59.58 & 13.09 \\
        $\lambda_{Imp}=1$, $\lambda_{Inter}=0.1$   & 71.81 & 69.13 & 56.38 & 15.43 \\
        $\lambda_{Imp}=100$, $\lambda_{Inter}=100$ & 63.01 & 61.64 & 54.20 & 8.81 \\
        \bottomrule
    \end{tabular}}
    \caption{Ablation study of $\lambda_{Imp}$ and $\lambda_{Inter}$. The choice of \(\lambda\) affects both prediction accuracy and model interpretability.}
    \label{tab:ablation}
\end{table}

%% file: figures/Sensitivity.tex
\begin{figure}[H]
    \centering
    \includegraphics[width=\linewidth]{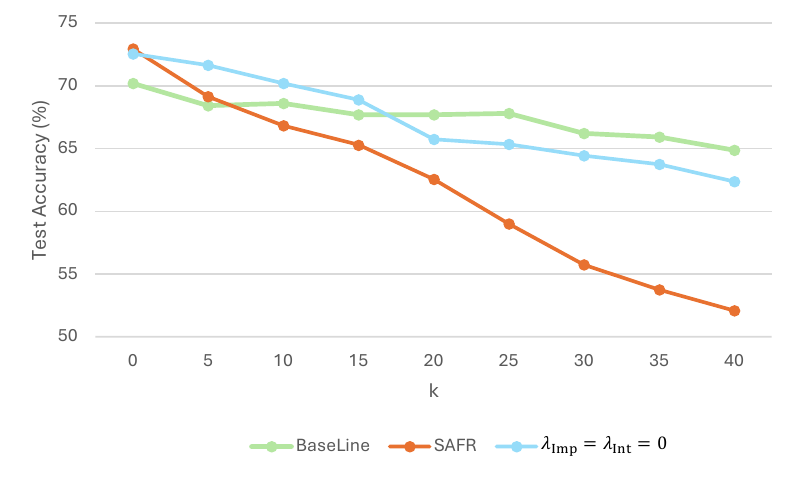}
    \caption{Sensitivity to $k$ Selection. As tokens are gradually removed, accuracy declines consistently.}
    \label{fig:sensitivity}
\end{figure}

%% file: tables/Capacity.tex
\begin{table}[ht]
\centering
\resizebox{\columnwidth}{!}{
\begin{tabular}{lcc}
\toprule
Average Capacity               & SST-2 & IMDB \\
\midrule
All Tokens       & 0.2981 & 0.1403 \\
Important Tokens (top 30\%) & \textbf{0.5745} & \textbf{0.2035} \\
Less Important Tokens (the rest 70\%) & 0.1794 & 0.1132 \\
\bottomrule
\end{tabular}
}
\caption{Average capacity metric for SST-2 and IMDB datasets. The metric reveals that important tokens exhibit significantly higher capacity scores compared to the overall average capacity.}
\label{tab:capacity}
\end{table}

%% file: sections/RelatedWork.tex
\section{Related Work}

Superposition in neural networks has gained attention, with foundational work by~\cite{arora-etal-2018-linear,Goh_2016}. \citet{olah2020zoom} developed this idea into the ``superposition hypothesis'' and initiated studies on mechanistic interpretability concerning polysemantic neurons and circuits. \citet{lecomte2024what} showed that polysemanticity can emerge incidentally through regularization and neural noise.

\citet{elhage2022toy} illustrated superposition in simplified networks, while subsequent works explored its theoretical, empirical, and applications~\cite{scherlis2022polysemanticity,henighan2023superposition,hanni2024mathematical,gurnee2023finding,marshall2024understanding,hanni2024mathematical,Katta2024OnIO,Chen2023DynamicalVB,chen-etal-2024-shot-named}.

Interpretability research includes works such as~\cite{Dreyer_2024_CVPR, Black2022InterpretingNN, Wang2023LearningFE}. Meanwhile, challenges in knowledge~\cite{Hu2024KnowledgeIS} and identifying universal feature spaces across models~\cite{lan2024sparseautoencodersrevealuniversal} mark promising directions for future research.

%% file: sections/Conclusion.tex
\section{Conclusion}
In this work, we introduced \ourmethod, an approach to enhance model interpretability by strategically regularizing feature superposition. Experiments on SST-2 and IMDB show that \ourmethod improves interpretability, as measured by our \eval metric, without compromising prediction performance.

Our method provides insights into the relationship between superposition and interpretability and offers a framework for visualizing neuron allocation. It contributes to mechanistic interpretability and suggests promising directions for extending the approach to larger models and wider applications.

%% file: sections/Limitation.tex
\section{Limitation}
This study has several limitations. First, experiments were conducted using a single-layer transformer model; future work should examine the scalability of \ourmethod with more complex architectures. Second, while focused on classification task, the applicability of \ourmethod to other NLP tasks—such as natural language inference, question answering, and text generation—remains unexplored. Third, there is a need for more comprehensive and standardized evaluation metrics to assess \ourmethod effectively. Finally, \ourmethod does not fully elucidate the causal mechanisms behind the model’s decision-making process. Addressing these challenges offers valuable opportunities for future research.

%% file: sections/Ethics.tex
\section{Ethic Statements}
Our research focuses on understanding and controlling the inner workings of transformer models, without collecting or using any human data; no personal or sensitive information is handled in this study. All datasets used in this work are public.

%% file: sections/Acknowledgment.tex
\section*{Acknowledgements}

We thank the anonymous reviewers for their valuable comments. We thank the Chili Lab at Rice for helpful discussions and suggestions throughout this research.

%% file: sections/Vmask.tex
\section{VMASK Introduction}
\label{vmask}
VMASK~\cite{chen-ji-2020-learning} is a variational word mask layer that is inserted into a neural text classifier and trained with the model. It learns to limit the flow of globally irrelevant or noisy word-level feature information to subsequent network layers, thus forcing the model to focus on the important features for prediction.

%% file: sections/Statistics.tex
\section{Statistics}
\label{statistics}
This section provides the statistical summaries of the datasets and the model.
\input{tables/Data}
\input{tables/Model}

%% file: tables/Data.tex
\begin{table}[H]
  \centering
  \resizebox{0.35\textwidth}{!}{%
  \begin{tabular}{lccc}
    \toprule
    \textbf{Datasets} & \textbf{\#Train} & \textbf{\#Dev} & \textbf{\#Test}\\
    \midrule
    SST2               & 6244 & 825 & 1749\\
    IMDB           & 20k & 5k & 25k\\
    \bottomrule
  \end{tabular}}
  \caption{Summary statistics for the datasets, where \# counts the number of examples in the train/dev/test sets.}
  \label{tab:data_statistics}
\end{table}

%% file: tables/Model.tex
\begin{table}[H]
  \centering
  \resizebox{0.35\textwidth}{!}{%
  \begin{tabular}{lccc}
    \toprule
    \textbf{Layer} & \textbf{Dimension}\\
    \midrule
    Embedding               & (Input Dimension, 256)\\
    FC1           & (256, 1024)\\
    FC2           & (1024, 256)\\
    \bottomrule
  \end{tabular}}
  \caption{Summary statistics for the model.}
  \label{tab:model_statistics}
\end{table}

%% file: sections/Visualization.tex
\section{Neuron Allocation Accross Layers}
\label{Neuron Allocation Accross Layers}
This section presents the observations regarding neuron allocation across the various layers, as visualized in Figure~\ref{fig:capacity-1} and \ref{fig:interference-1}.

\input{figures/Visualization}

%% file: figures/Visualization.tex
\begin{figure*}
    \centering
      \includegraphics[width=0.42\linewidth]{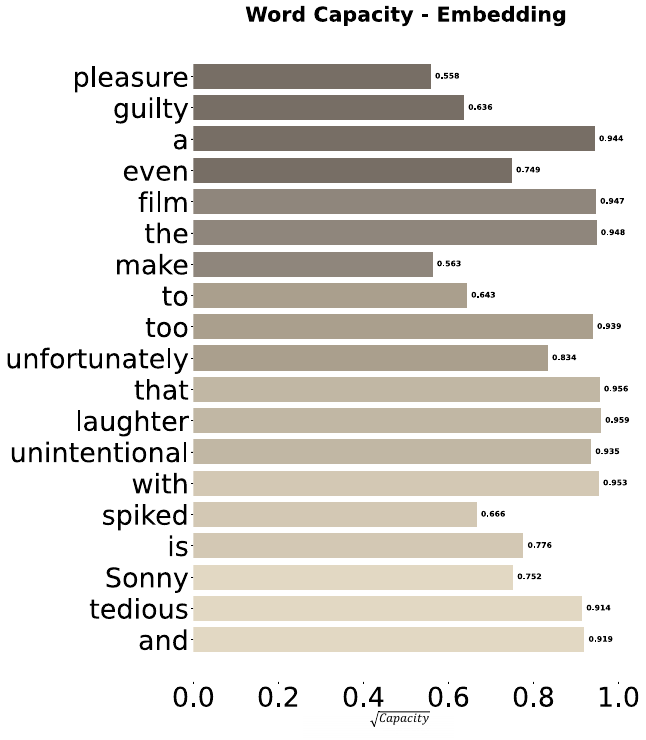}
    \includegraphics[width=0.42\linewidth]{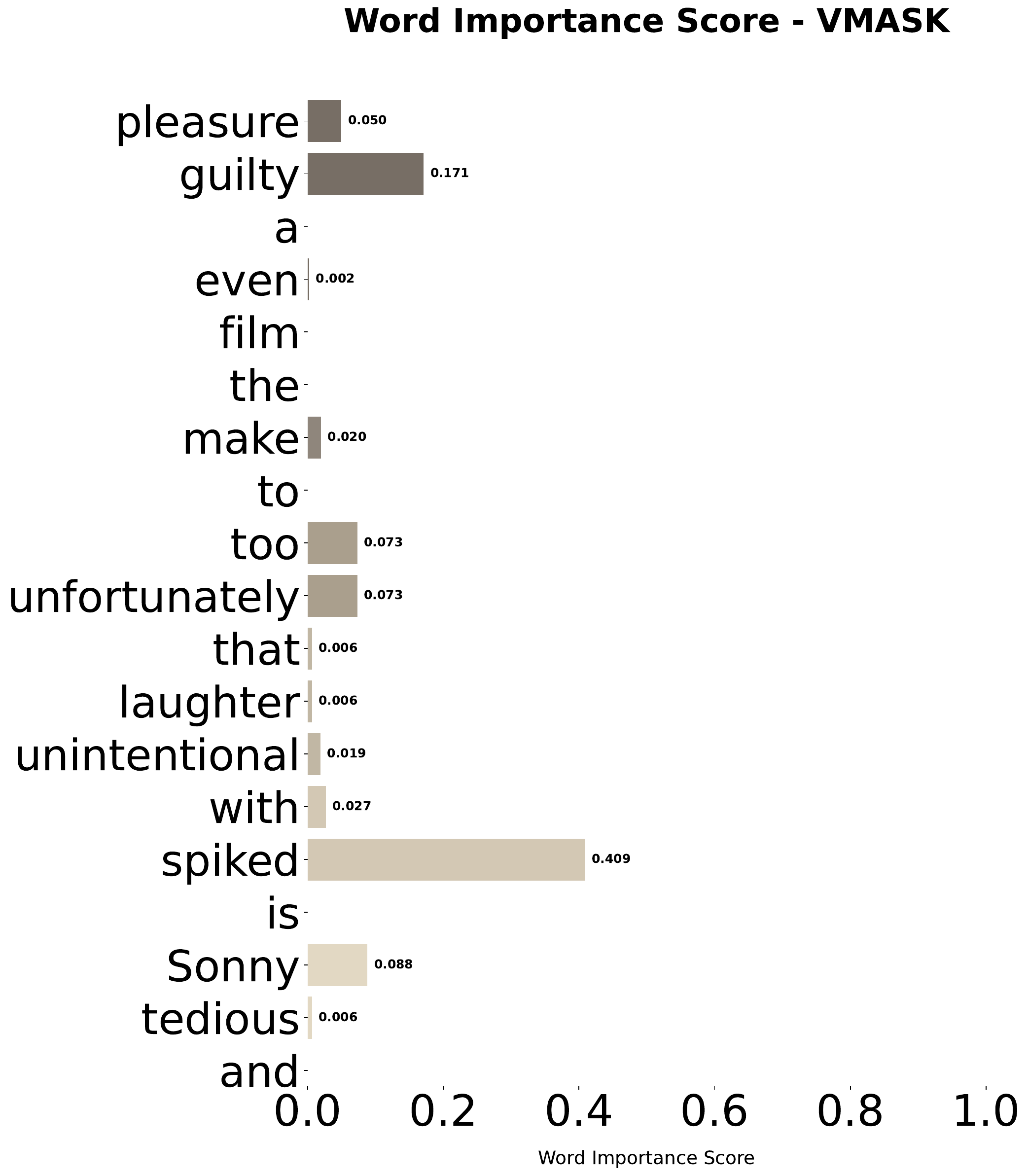}
      \includegraphics[width=0.42\linewidth]{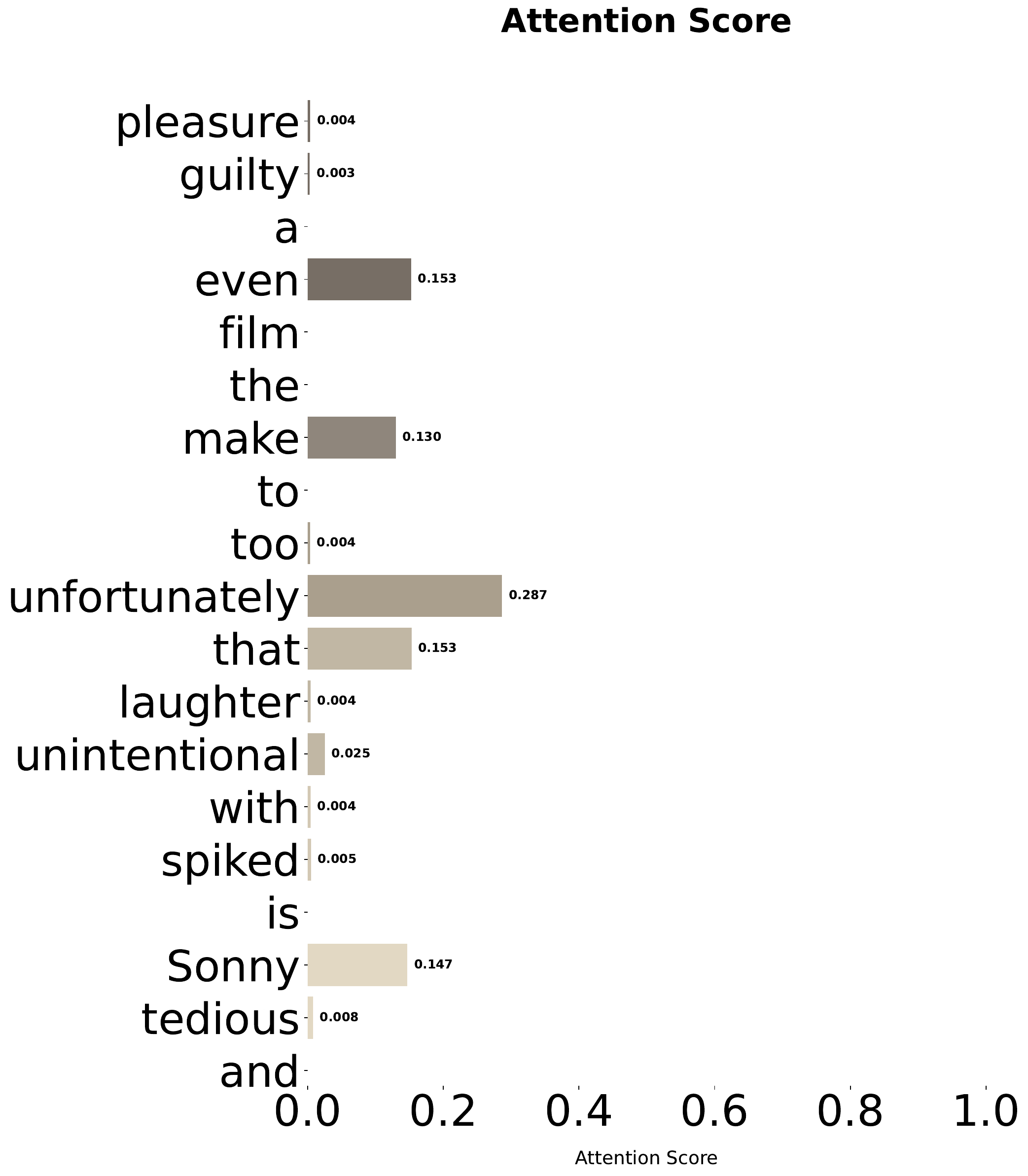}
      \includegraphics[width=0.42\linewidth]{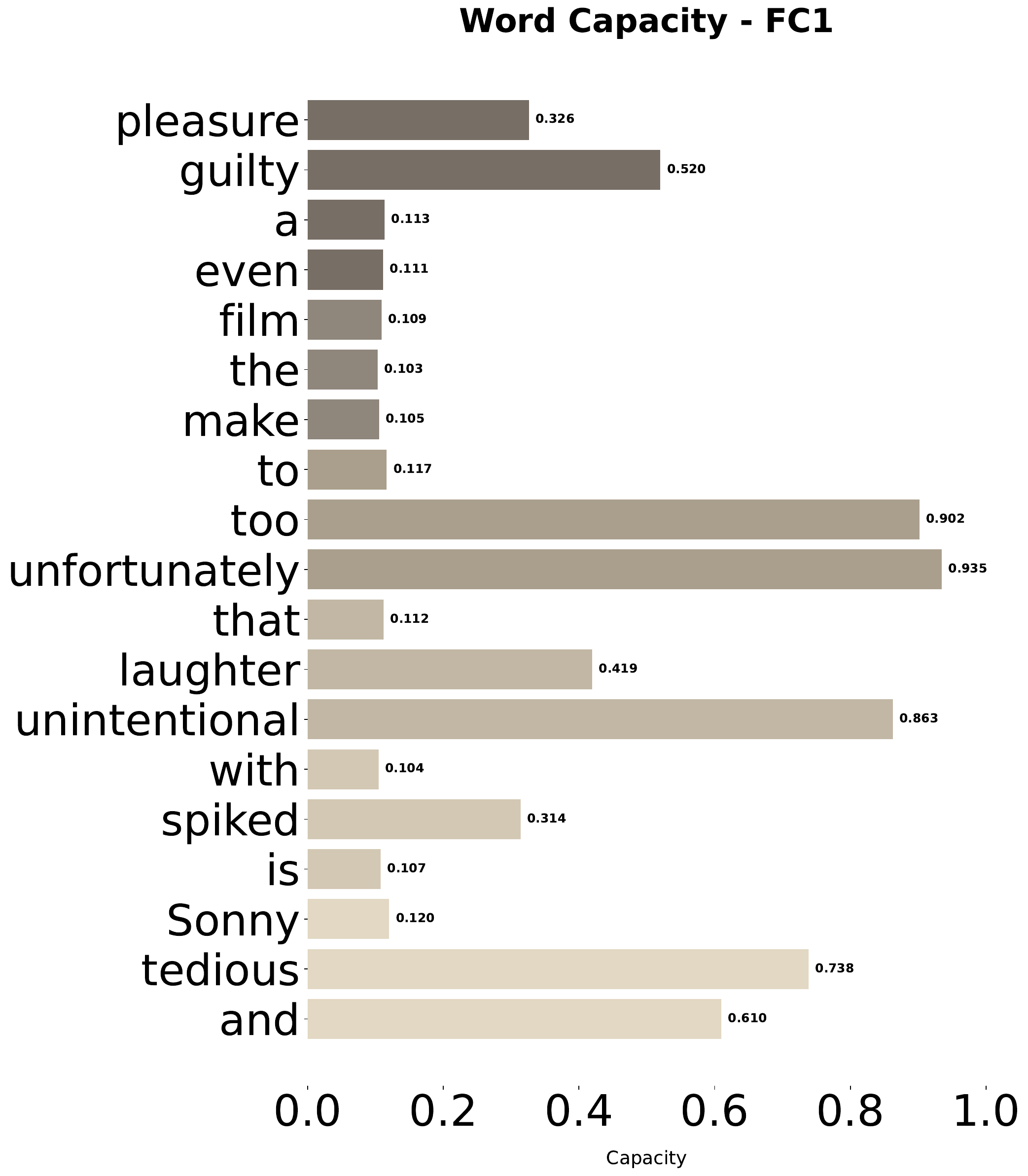}
      \includegraphics[width=0.42\linewidth]{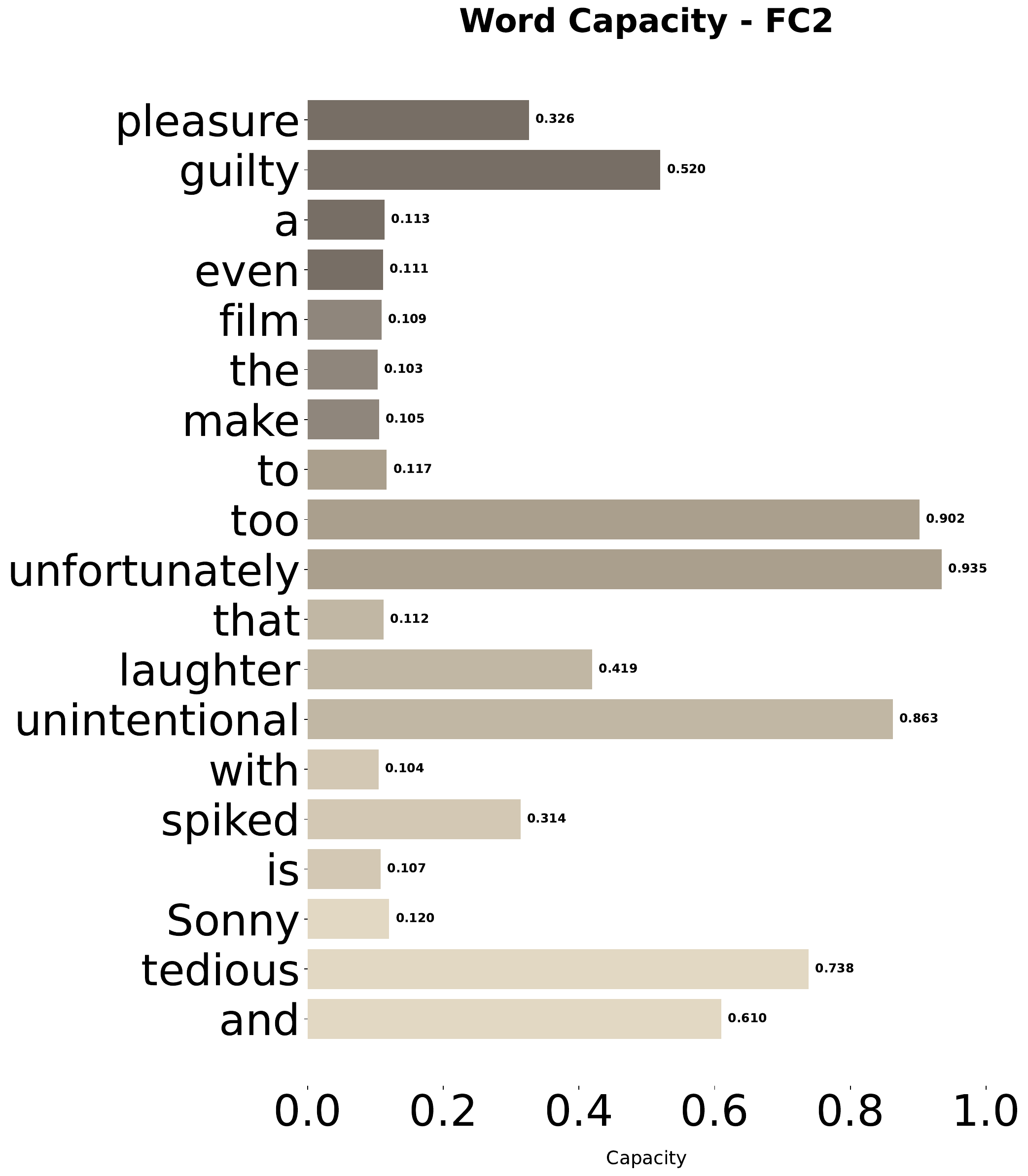}
    \caption{Cross Layers Output: Capacity. VMASK layer uses the importance scores it detects, while the attention layer uses normalized attention scores. The original sentence is ``Preposterous and tedious, Sonny is spiked with unintentional laughter that, unfortunately, occurs too infrequently to make the film even a guilty pleasure.''(negative)}
\label{fig:capacity-1}
\end{figure*}

\begin{figure*}
    \centering
  \includegraphics[width=0.45\linewidth]{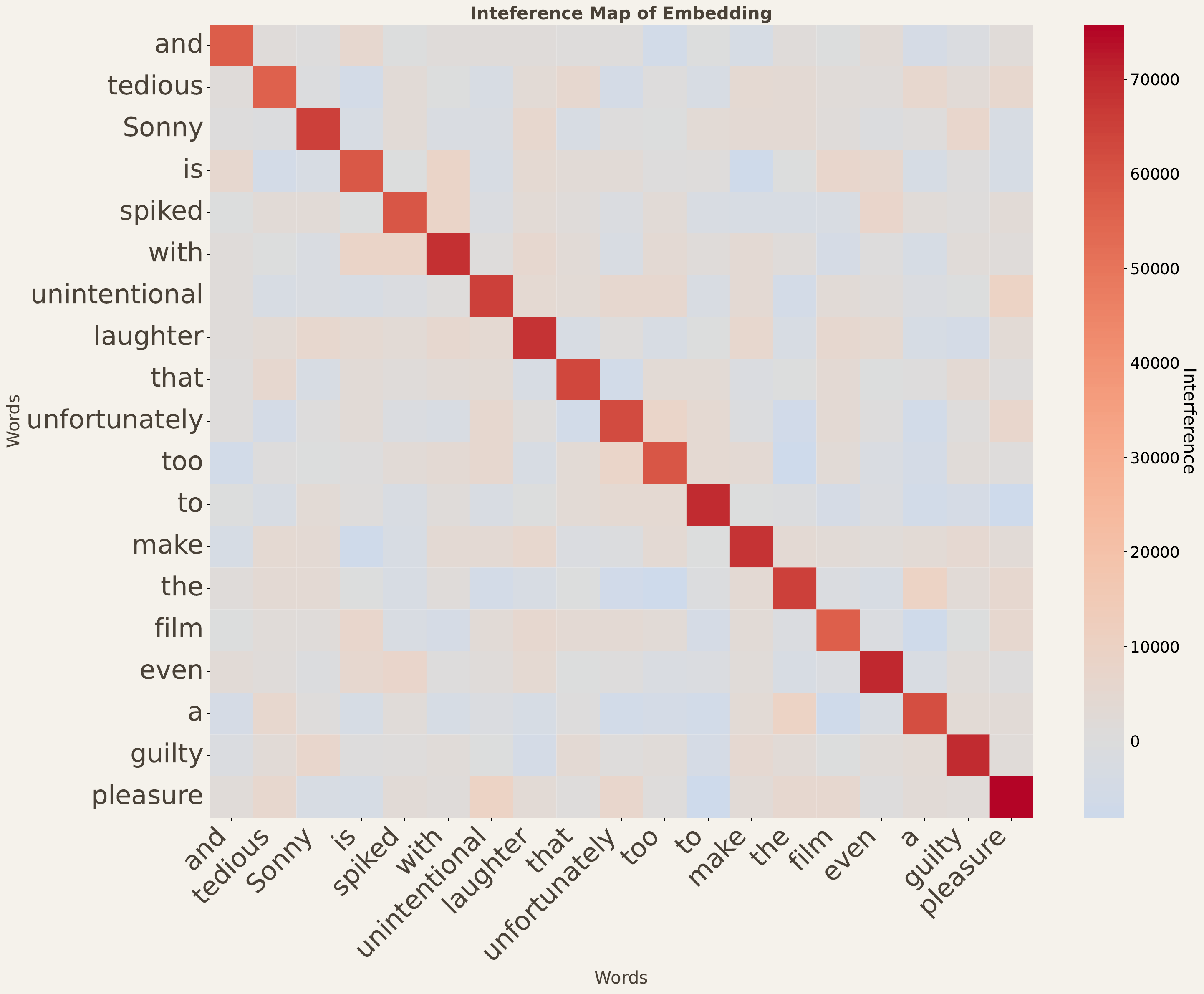}
  \includegraphics[width=0.45\linewidth]{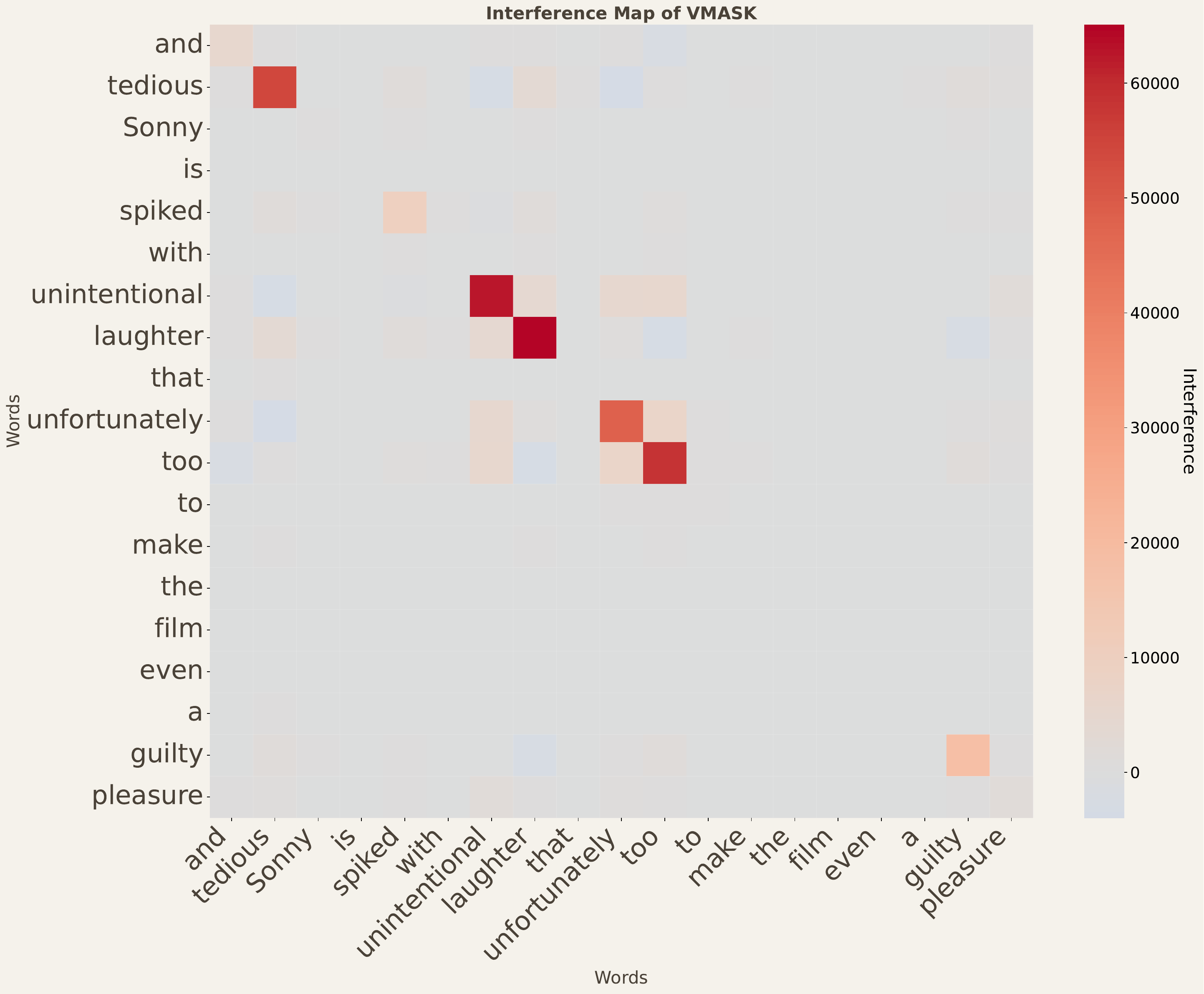}
  \includegraphics[width=0.45\linewidth]{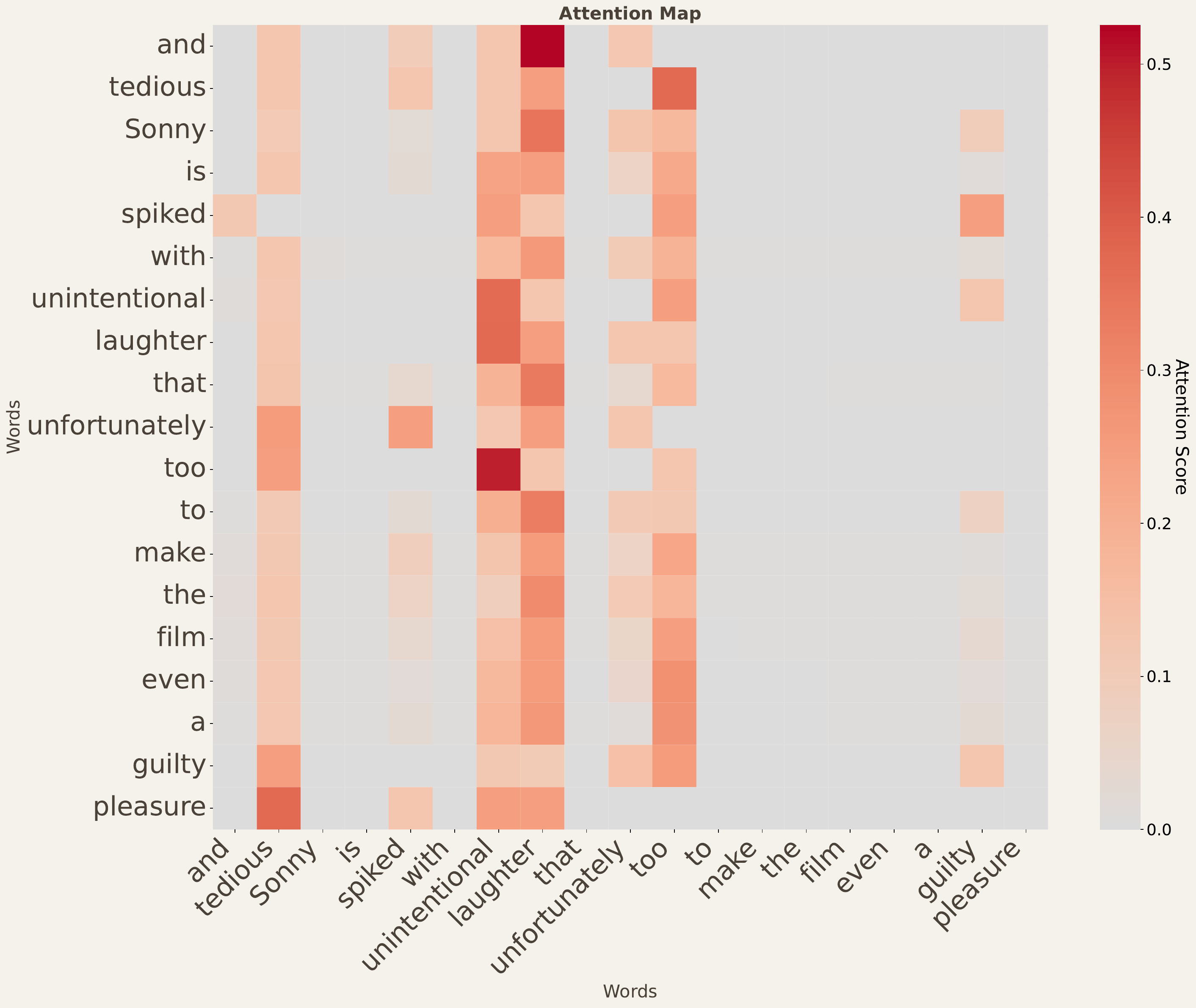}
  \includegraphics[width=0.45\linewidth]{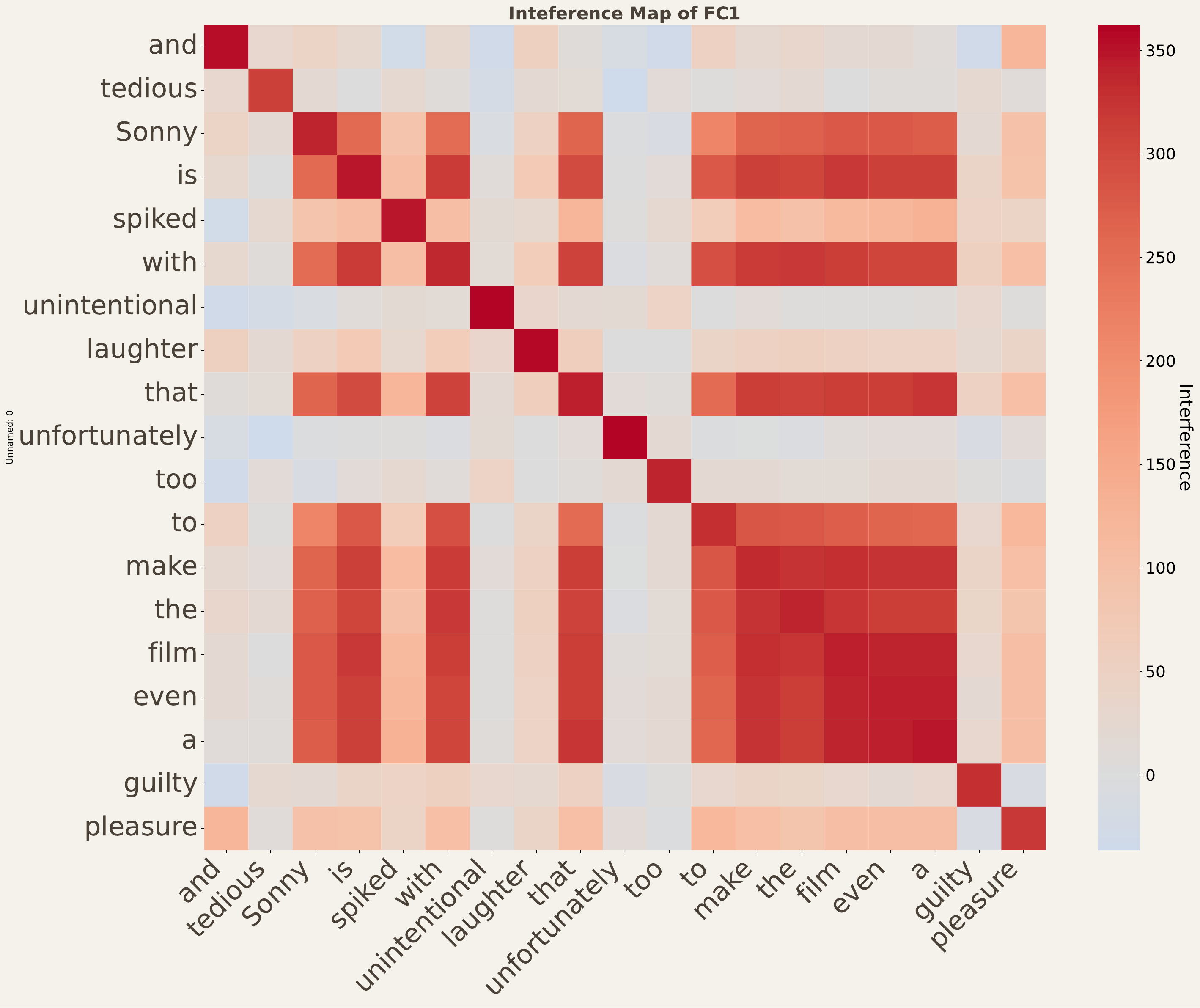}
  \includegraphics[width=0.45\linewidth]{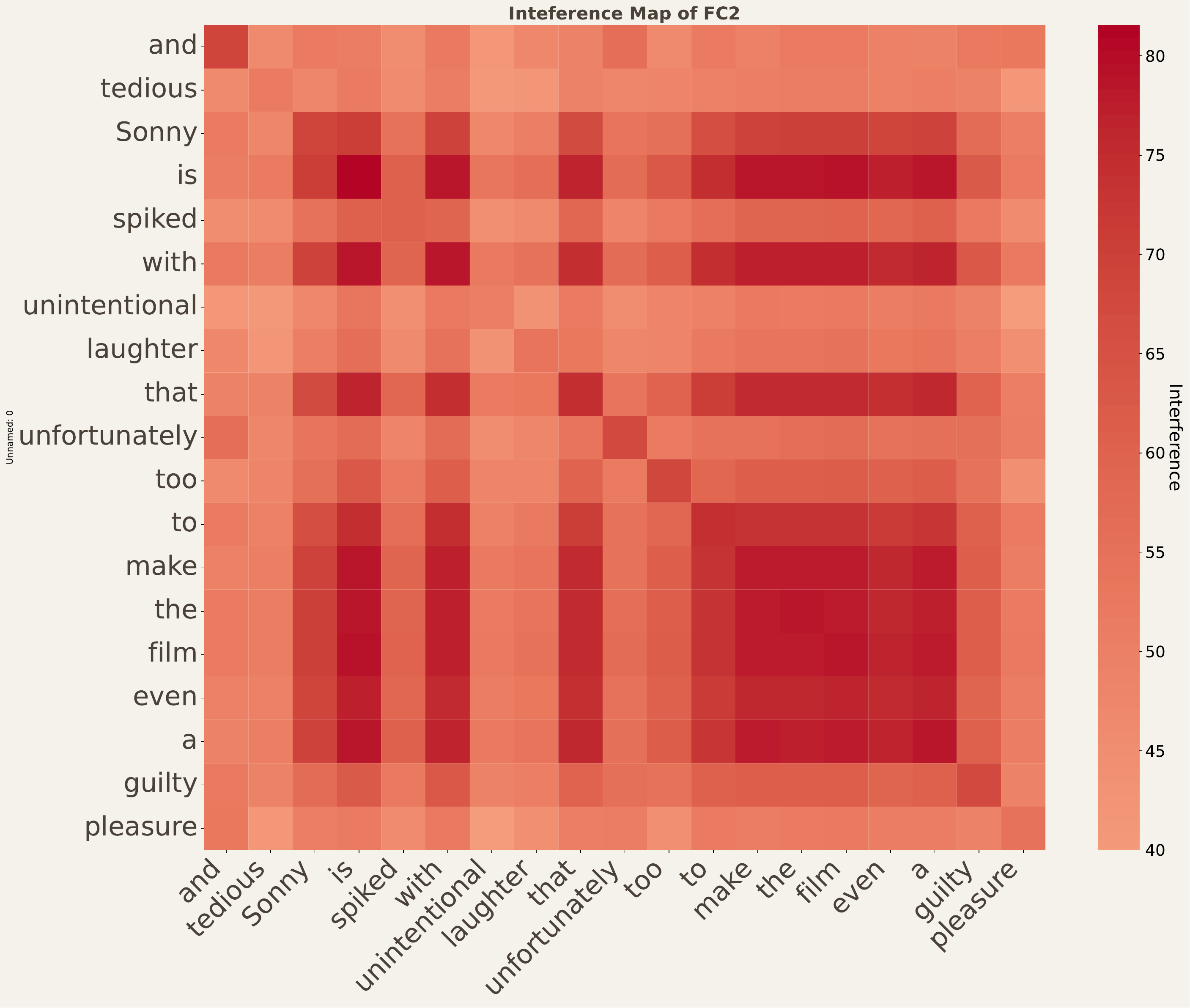}
\caption{Cross Layers Output: Interference. The attention layer uses the attention weight matrix. The original sentence is ``Preposterous and tedious, Sonny is spiked with unintentional laughter that, unfortunately, occurs too infrequently to make the film even a guilty pleasure.''(negative)}
\label{fig:interference-1}
\end{figure*}

%% file: sections/Ablation.tex
\input{tables/Ablation_full}
\section{Regularization Hyperparameter Tuning}
\label{Regularization Hyperparameter Tuning}
This section presents the results of hyperparameter tuning for $\lambda_{Imp}$ and $\lambda_{Inter}$, as summarized in Table~\ref{tab:ablation_full}.

%% file: tables/Ablation_full.tex
\begin{table*}
    \centering
    \renewcommand{\arraystretch}{0.8}
    \resizebox{\textwidth}{!}{%
    \begin{tabular}{lcccccccc}
        \toprule
        & \multicolumn{4}{c}{SST-2} & \multicolumn{4}{c}{IMDB} \\
        \cmidrule(lr){2-5} \cmidrule(lr){6-9}
        Model & $\text{Acc}_{S} (\%)$ & $\text{Acc}_{\tilde{S}^{(r)}} (\%)$ & $\text{Acc}_{\tilde{S}^{(k)}} (\%)$ & \eval & $\text{Acc}_{S} (\%)$ & $\text{Acc}_{\tilde{S}^{(r)}} (\%)$ & $\text{Acc}_{\tilde{S}^{(k)}} (\%)$ & \eval \\
        \midrule
        Baseline & 70.21 & 67.12 & 66.21 & 4.00 & 80.14 & 77.11 & 76.60 & 3.54 \\
        $\lambda_{Imp}=0$, $\lambda_{Inter}=0$    & 72.56 & 69.47 & 64.44 & 8.12  & 78.43 & 76.10 & 73.56 & 4.87 \\
        $\lambda_{Imp}=0$, $\lambda_{Inter}=0.01$  & 72.84 & 70.27 & 62.78 & 10.06 & 81.16 & 78.20 & 74.61 & 6.55 \\
        $\lambda_{Imp}=0$, $\lambda_{Inter}=1$     & 73.01 & 69.07 & 60.32 & 12.69 & 80.44 & 76.89 & 71.84 & 8.60 \\
        $\lambda_{Imp}=0$, $\lambda_{Inter}=100$   & 59.35 & 57.75 & 56.38 & 2.97 & 77.83 & 74.90 & 67.64 & 10.19 \\
        $\lambda_{Imp}=0.01$, $\lambda_{Inter}=0$  & 72.67 & 69.87 & 61.41 & 11.26 & 80.00 & 77.02 & 70.98 & 9.02 \\
        $\lambda_{Imp}=0.01$, $\lambda_{Inter}=0.01$ & 73.07 & 71.07 & 58.49 & 14.58 & 80.30 & 77.42 & 64.58 & 15.72 \\
        $\lambda_{Imp}=0.1$, $\lambda_{Inter}=0.1$ & 72.96 & 70.61 & 55.75 & 17.21 & 79.24 & 75.96 & 53.99 & 25.25 \\
        $\lambda_{Imp}=0.1$, $\lambda_{Inter}=1$   & 72.67 & 69.18 & 59.58 & 13.09 & 78.45 & 75.05 & 49.97 & 28.48 \\
        $\lambda_{Imp}=1$, $\lambda_{Inter}=0$     & 71.30 & 66.96 & 55.06 & 16.24 & 74.95 & 72.30 & 50.07 & 24.88 \\
        $\lambda_{Imp}=1$, $\lambda_{Inter}=0.1$   & 71.81 & 69.13 & 56.38 & 15.43 & 74.89 & 71.72 & 50.08 & 24.81 \\
        $\lambda_{Imp}=1$, $\lambda_{Inter}=1$     & 71.93 & 70.67 & 56.78 & 15.15 & 74.78 & 71.44 & 53.94 & 20.84 \\
        $\lambda_{Imp}=100$, $\lambda_{Inter}=0$   & 64.67 & 63.75 & 57.92 & 6.75  & 51.53 & 51.20 & 50.00 & 1.53 \\
        $\lambda_{Imp}=100$, $\lambda_{Inter}=100$ & 63.01 & 61.64 & 54.20 & 8.81 & 53.48 & 53.33 & 50.00 & 3.48 \\
        \bottomrule
    \end{tabular}}
    \caption{Ablation Study on SST-2 and IMDB datasets. The choice of \(\lambda\) affects both prediction accuracy and model interpretability.}
    \label{tab:ablation_full}
\end{table*}